\documentclass[10pt,twocolumn,letterpaper]{article}
\usepackage[accsupp]{axessibility}
\usepackage[table]{xcolor}
\usepackage{cvpr}              

\newcommand{\drule}{\specialrule{0.2pt}{1pt}{1pt}%
            \specialrule{0.2pt}{0pt}{\belowrulesep}%
            }

\definecolor{cvprblue}{rgb}{0.21,0.49,0.74}
\usepackage[pagebackref,breaklinks,colorlinks,allcolors=cvprblue]{hyperref}
\def\paperID{42536} 
\def\confName{CVPR}
\def\confYear{2026}

\title{AHS: Adaptive Head Synthesis via Synthetic Data Augmentations}

\author{Taewoong Kang$^{1}$\thanks{Equal contribution} \quad Hyojin Jang$^{1}$\footnotemark[1] \quad Sohyun Jeong$^{1}$\footnotemark[1] \quad Seunggi Moon$^{2}$ \\
\quad Gihwi Kim$^{3}$ \quad Hoon Jin Jung$^{3}$  \quad Jaegul Choo$^{1}$ \\
$^{1}$KAIST $^{2}$ Korea University $^{3}$ FLIPTION \\
\texttt{\{keh0t0, wkdgywlsrud, jsh0212, jchoo\}@kaist.ac.kr} }

\begin{document}

\twocolumn[{%
\renewcommand\twocolumn[1][]{#1}%
\maketitle
\begin{center}
    \centering
    \captionsetup{type=figure}
\includegraphics[width=1.0\linewidth]{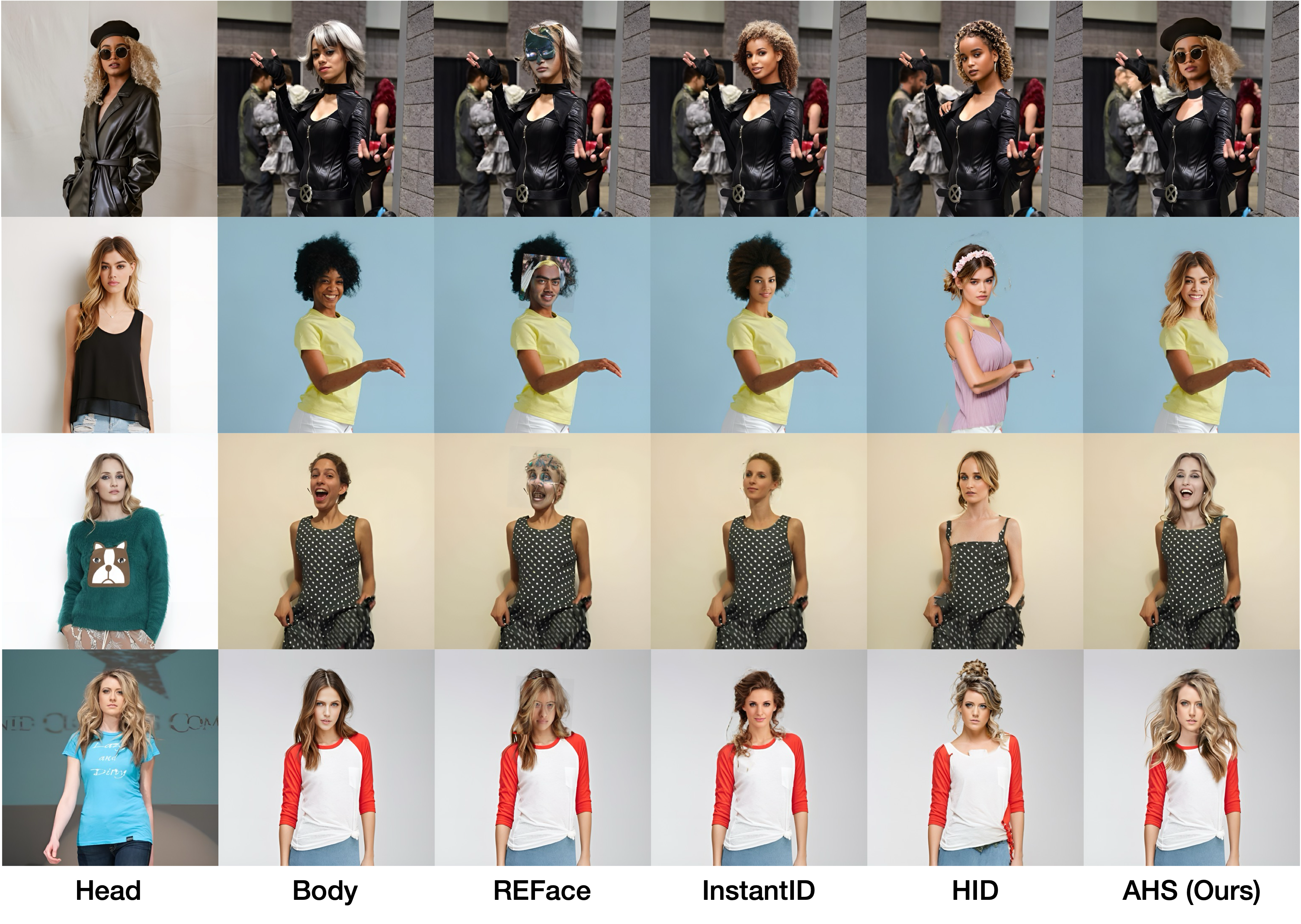}
    \captionof{figure}{\textbf{Head-swapped results comparison among the baselines.} Our model outperforms on preserving identity, hairstyle and accessaries while reeancting target body image's expression and head pose.
    }
    \label{fig:teaser}
\end{center}
}]

\begin{abstract}
Recent digital media advancements have created increasing demands for sophisticated portrait manipulation techniques, particularly head swapping, where one's head is seamlessly integrated with another's body. However, current approaches predominantly rely on face-centered cropped data with limited view angles, significantly restricting their real-world applicability. They struggle with diverse head expressions, varying hairstyles, and natural blending beyond facial regions. To address these limitations, we propose Adaptive Head Synthesis (AHS), which effectively handles full upper-body images with varied head poses and expressions. AHS incorporates a novel head reenacted synthetic data augmentation strategy to overcome self-supervised training constraints, enhancing generalization across diverse facial expressions and orientations without requiring paired training data. Comprehensive experiments demonstrate that AHS achieves superior performance in challenging real-world scenarios, producing visually coherent results that preserve identity and expression fidelity across various head orientations and hairstyles. Notably, AHS shows exceptional robustness in maintaining facial identity while drastic expression changes and faithfully preserving accessories while significant head pose variations.
\end{abstract}


\section{Introduction}
Head swapping is a challenging task that seamlessly integrates a head of a source image with a body of a target image, while reenacting the head orientation and expression of the target image.
Given its potential impact on industries such as fashion design, virtual character customization, and digital marketing, exploring this task holds considerable research value.
Despite its potential, head swapping remains relatively underexplored due to its inherent challenges. One major challenge is the lack of ground truth data for head swapping. As a result, models only could rely on self-supervised training, which significantly weakens their generalization capabilities. Specifically, models trained solely on self-reconstruction often struggle to capture variations in facial expressions and head orientations, limiting their effectiveness in real-world applications. 
Another key challenge arises from the high variability in hair length and style, requiring the model to consider a broader spatial region. This makes head swapping more difficult than face swapping~\cite{chen2020simswap, infoswap, selfswapper}, a similar task that transfers the facial identity of a source to a target while preserving the target’s non-identity-related attributes, as it focuses exclusively on the facial region.

\begin{figure}[t]
  \centering
   \includegraphics[width=0.9\linewidth]{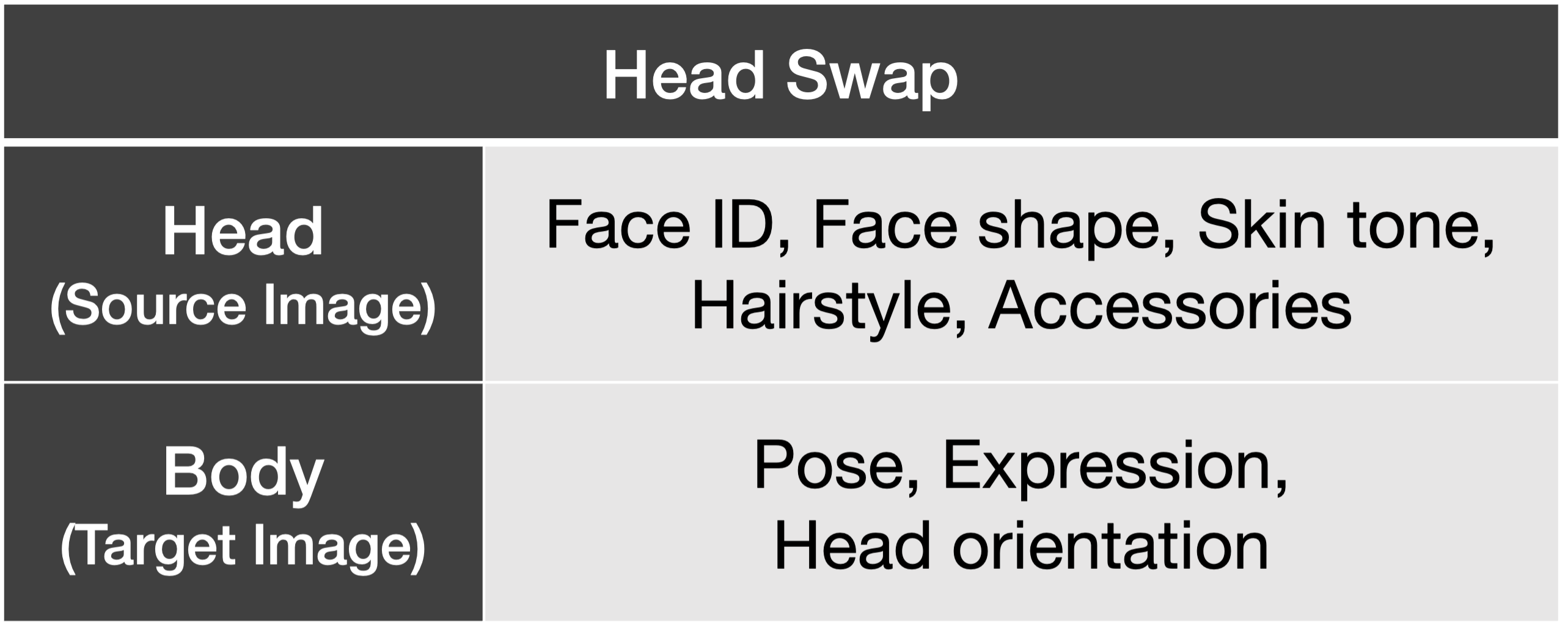}
   \caption{\textbf{Problem definition of head swapping.} The first row indicates the portion of the head from the source image that needs to be transferred, while the second row indicates the portion of the head in the target image that should be preserved.}
   \label{fig:def}
\end{figure}

Existing head swapping approaches have attempted to address these challenges, but significant limitations still hinder their effectiveness in real-world applications. One approach restricts the editing region to the cropped face area to simplify the task by reducing spatial complexity~\cite{deepfacelab, reface, faceX}. However, this severely impacts practical applicability. Since head swapping inherently involves variations in hair shape, length, and overall head orientation, confining modifications to only the facial region prevents seamless integration. As a result, this approach falls short in applications that require full head synthesis. Another approach leverages few-shot training techniques~\cite{heser} to reenact the head orientation from the target image. It often necessitates complex preprocessing, as it typically relies on video data to extract training samples. Additionally, it commonly consists of two separate models, one for reenactment and another for blending, further increasing computational complexity. Moreover, it is generally less effective than zero-shot methods, which need no prior data on specific individuals. Given these drawbacks, there is a growing need for more robust head swapping methods that can operate in a zero-shot setting using a single model while effectively handling variations in head orientation, hair structure, and expression without requiring extensive preprocessing or specialized training data. HID~\cite{ours} tackles these issues and enables zero-shot head swapping.
However, it still has difficulty capturing facial attributes such as expressions and relies solely on ControlNet~\cite{controlnet}-OpenPose during inference to manage pose and head orientation, as its training phase lacks mechanisms for controlling these aspects. For a more natural and practical outcome, it is crucial to transfer the face ID, facial shape, skin tone, accessories, and hairstyle of the source image, while preserving the pose, expression, and head orientation of the target image, as shown in \cref{fig:def}.

Therefore, we propose Adaptive Head Synthesis (AHS), which designed to effectively handle diverse facial orientations, expressions, and hairstyles in challenging settings of full upper body images. To overcome the limitations of self-reconstruction training caused by the absence of ground-truth data, we introduce a synthetic data augmentation strategy using a state-of-the-art animatable head avatar model~\cite{GAGAvatar}. This augmentation enhances the zero-shot adaptability, making AHS a robust and practical solution for real-world head-swapping applications. Furthermore, we achieve precise control over expression, head orientation, and pose by simply merging two conditioning images, a densepose map~\cite{densepose} and a normal map.

Our main contributions are as follows:

\begin{itemize}
    \item We propose AHS, a novel approach that enables effective and high-quality head swapping with a single model on challenging datasets, while considering head orientation alignment and expression reenactment.

    \item We present a synthetic data augmentation strategy using a head reenactment model, which mitigates the limitations of self-supervised training and enhances generalization capabilities across diverse facial expressions, head orientations, and hairstyles.

    \item Through extensive experiments, we demonstrate that AHS achieves state-of-the-art head swapping performance in complex real-world scenarios.
\end{itemize}

\section{Related Work}
\subsection{Head Swap}
Although research on head swapping remains relatively underexplored, existing studies~\cite{deepfacelab, faceX, reface, hsdiffusion, heser, ghost} share a common limitation: they predominantly rely on face-centered cropped datasets, which primarily consist of front-facing views. This dataset constraint severely limits the diversity of head orientations and seamless head-body integration, which are crucial for real-world applications. More details for these works are provided in \cref{Appen:Related}.

To overcome these dataset constraints, for example, in real-world cases where subjects have extremely long hair, recent work~\cite{ours} leverages a real-world dataset and introduces a strategy to improve generalization by injecting hair and face ID information through text embeddings.
However, this approach suffers from artifact generation and a decline in identity similarity, because it relies on embedding-level injection rather than feature-level injection. Additionally, it does not explicitly model facial expressions, limiting its effectiveness in capturing natural variations. 
These limitations highlight the need for a more robust head swapping approach that can generalize effectively to diverse head orientations, expressions, and hairstyles without being constrained by face-centered cropped datasets. 

\subsection{Diffusion-based Image Editing}
Diffusion models~\cite{sd, ho2020denoising, song2020denoising} have achieved substantial advances in text-to-image synthesis~\cite{imagic, ramesh2022hierarchical, rombach2022high, saharia2022photorealistic}, drawing widespread interest in recent years. These developments have been propelled by the emergence of extensive and high-quality text-image datasets~\cite{changpinyo2021conceptual, schuhmann2022laion}, ongoing enhancements in foundational models~\cite{chen2023pixart, peebles2023scalable}, improvements in conditioning encoders, and the introduction of sophisticated control mechanisms~\cite{mou2024t2i, li2023gligen, ipadapter, controlnet, ip2p}. Recent research has increasingly focused on leveraging diffusion models for image editing under diverse conditions. Such methods include pose-guided generation~\cite{animate_any, tcan}, reference-based editing ~\cite{paintbyex, anydoor}, multi-view face synthesis ~\cite{joker}, and multi-conditional image manipulation ~\cite{controlnet, reference}, which incorporate depth and edge maps. In addition, diffusion-based techniques have been applied to various inpainting tasks using reference images, such as object inpainting~\cite{paintbyex, anydoor}, hairstyle transformation~\cite{hairfusion}, and virtual clothing synthesis~\cite{stableviton, idm, zhu2023tryondiffusion}. Nevertheless, head inpainting poses distinct challenges compared to other inpainting tasks such as clothing synthesis. The wide range of hairstyles, facial orientations, and expressions leads to more complex inpainting regions, requiring precise control and making head manipulation particularly demanding. To bridge this gap, our work extends diffusion models to the head-swapping task, which involves transferring the head from a reference image onto a target body image. Our proposed approach effectively manages the complexities associated with head swapping while maintaining realistic and seamless integration.

\section{Method}

\begin{figure*}
  \centering
   \includegraphics[width=1.0\linewidth]{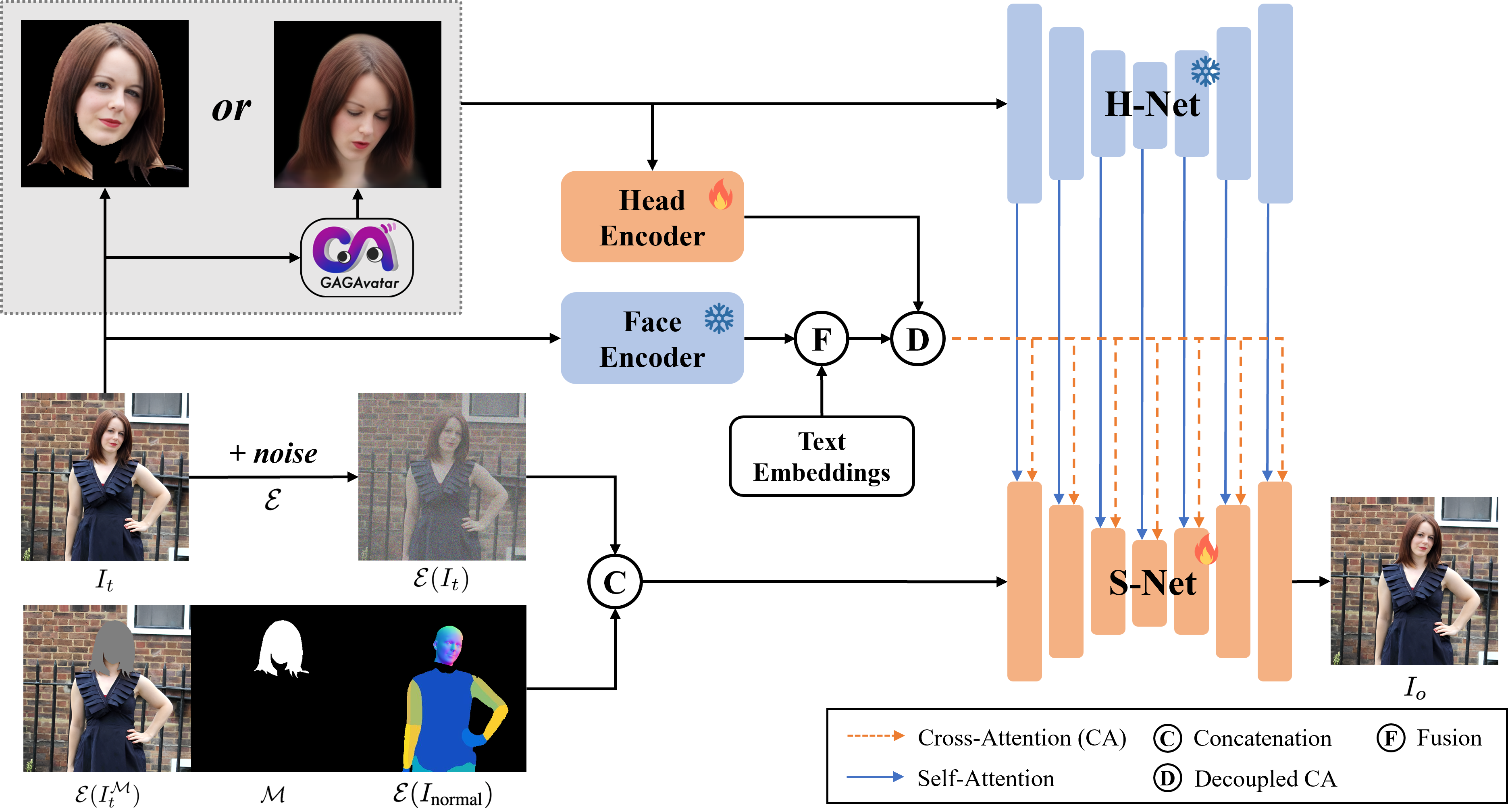}
   \caption{\textbf{Overview of training AHS.} Our model encodes identity using dedicated Head and Face Encoders, while H-Net preserves fine-grained head details. To prevent reconstruction artifacts and improve robustness, the training process is enhanced with GAGAvatar, which generates augmented data with diverse head poses and expressions.}
   \label{fig:framework}
\end{figure*}


This section presents our proposed framework for effective head swapping and reenactment, which comprises a novel data augmentation strategy and a specialized network architecture. 
In \cref{subsec:model}, we introduce our model architecture, designed to preserve identity and accessories with high fidelity. 
Following this, in \cref{subsec:data}, we detail our data augmentation strategy, which relies on a synthetic dataset of randomly reenacted heads. 
Finally, we describe effective inference in \cref{subsec:infer} utilizing our trained model.

\subsection{Model Architecture}
\label{subsec:model}
Let $I_s$ and $I_t$ be the source and target images, respectively. Our goal is to generate an output image $I_o$ where the head from $I_s$ seamlessly replaces the head in $I_t$. The generated head must preserve the identity from $I_s$ while matching the pose and expression of $I_t$. 

To achieve this alignment, we guide our model with $I_{\text{normal}}$. This is obtained by replacing the head region of the target's dense pose map with a normal map extracted via the state-of-the-art head reconstruction model, EMOCA~\cite{emoca}. This simple additional input provides explicit geometric cues, enabling effective reenactment of the target's head attributes.
The model takes the following inputs:
\begin{equation}\label{eq:ca}
    I_o = \Phi(\mathcal{E}(I_t), \mathcal{E}(I_t^\mathcal{M}), \mathcal{M}, \mathcal{E}(I_{\text{normal}})),
\end{equation}
where $\mathcal{E}$ represents a VAE encoder, $\mathcal{M}$ is the mask, $I_t^\mathcal{M}$ means masked $I_t$ and $\Phi$ denotes our full model.

To effectively integrate head identity, we employ both cross-attention and self-attention mechanisms within our primary U-Net, referred to as S-Net.
For cross-attention, we leverage the face encoder and head encoder inspired by Photomaker~\cite{photomaker} and IP-Adapter~\cite{ipadapter}. The face encoder extracts key-value matrices $K_f\in\mathbb{R}^{N\times d}$, $V_f\in\mathbb{R}^{N\times d}$ by fusing with text embeddings, while the head encoder computes $K_h\in\mathbb{R}^{N\times d'}$, $V_h\in\mathbb{R}^{N\times d'}$ from head embeddings $h$. 
The extracted information is incorporated into the model via the cross-attention layer:
\begin{equation}\label{eq:ca}
    \texttt{Attention}(Q, K_f, V_f) + \texttt{Attention}(Q, K_h, V_h).
\end{equation}
This enables the model to capture high-level semantic features, ensuring identity preservation while allowing flexible pose adaptation.
We further incorporate self-attention, following~\cite{idm}, to provide low-level features. Specifically, we extract key-value pairs $K_n\in\mathbb{R}^{N\times d''}$, $V_n\in\mathbb{R}^{N\times d''}$ from H-Net, the reference net, and concatenate them with those from S-Net. 
This allows the query to attend to both H-Net and S-Net features effectively referencing the source image features.
By combining self-attention for low-level features with cross-attention for high-level semantic control, our model effectively synthesizes head-swapped images that are both perceptually realistic and preserve intricate details. 
Moreover, the face and head encoders accelerate model convergence and compensate for cases where H-Net is not explicitly trained, ensuring stable performance even without extensive fine-tuning.

\subsection{Data Augmentation}
\label{subsec:data}
While our model architecture excels at transferring reference features, it faces challenges in pose alignment and photorealistic inpainting when trained for the head swapping task in a self-supervised manner.
For instance, in the case of in-the-wild images, the source and target images often exhibit significantly different head orientations and expressions, as they originate from different individuals.
However, this alignment issue cannot be fully addressed within the constraints of self-supervised training. 
To improve robustness and performance across diverse scenarios, we incorporate a strategic data augmentation.

The primary challenge lies in the inability of the model to effectively capture both head orientation and facial expression information. 
To address this challenge, we devise a simple yet effective approach by strategically augmenting the training dataset. 
Specifically, we leverage the state-of-the-art head reenactment model, GAGAvatar~\cite{GAGAvatar}. By altering the head orientation and facial expression of randomly selected images while minimally compromising the original identity information, we enable our model to generalize more effectively across various real-world pose pairs.
Through this augmentation, our model inherently learns head reenactment within a unified manner.

\begin{figure}[t]
  \centering
   \includegraphics[width=0.95\linewidth]{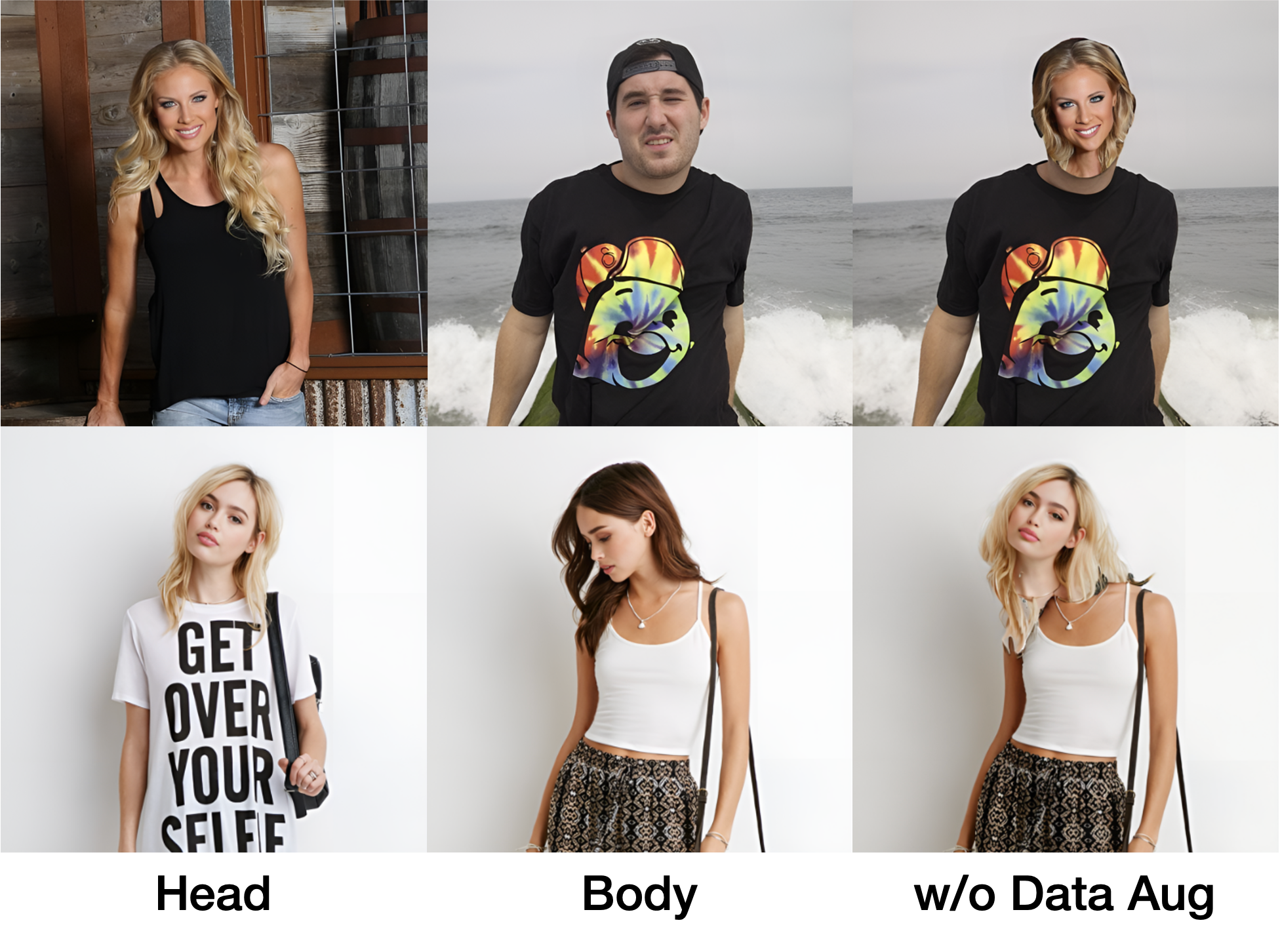}
   \caption{\textbf{Examples of Copy-and-paste artifacts.} These results are generated by the model that trained without applying our data augmentation methods, including head reenactment and masking. This copy-and-paste artifacts caused by not accounting for the pose and expression of the head in the target image.}  
   \label{fig:artifacts}
\end{figure}

\begin{figure}[t!]
  \centering
   \includegraphics[width=0.95\linewidth]{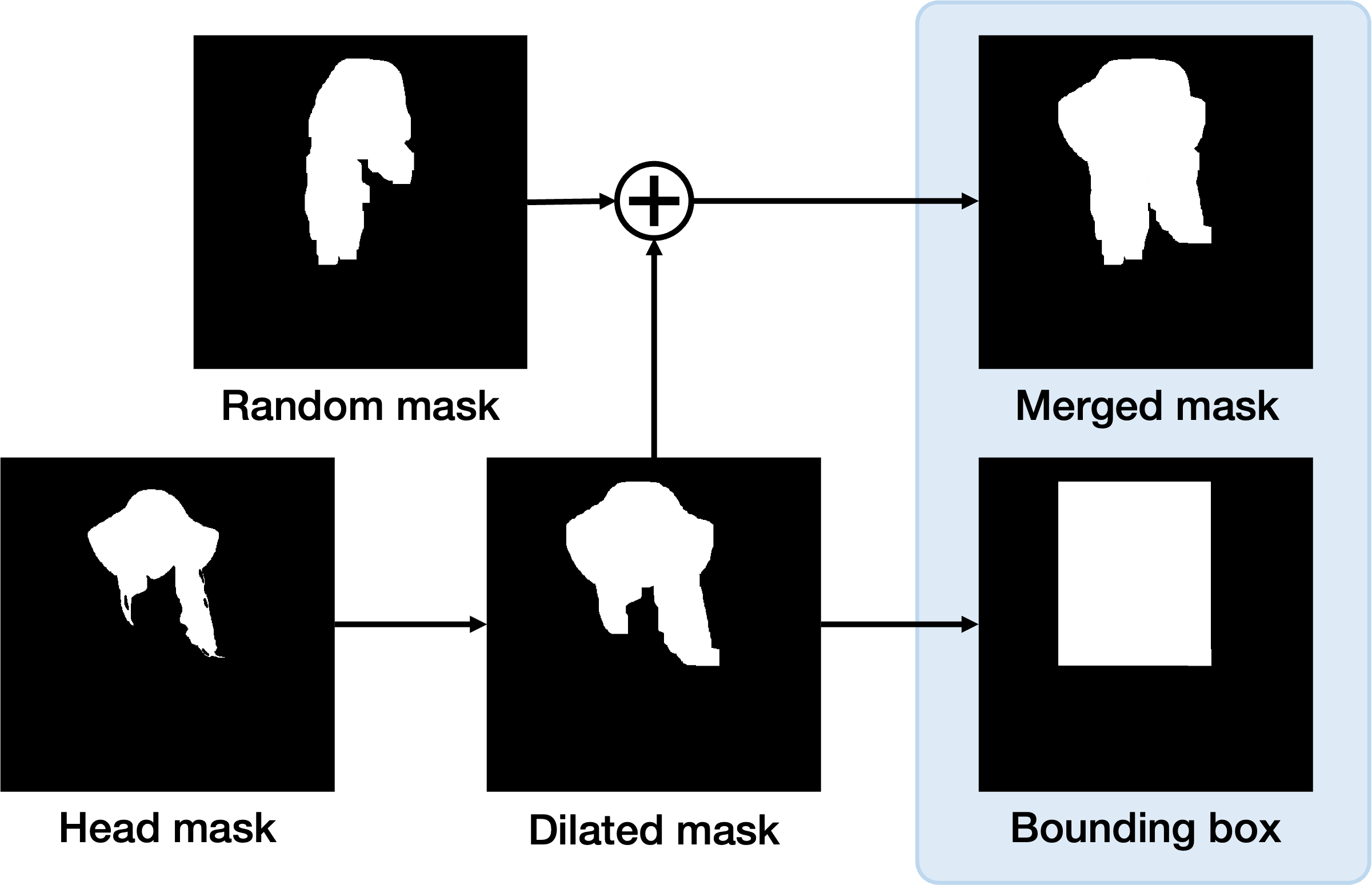}
   \caption{\textbf{Mask augmentation strategy.} Including dilation, widened bounding box creation, and merging with a random mask.}
    \label{fig:mask}
\end{figure}

Additionally, to achieve photorealistic head swapping, the heterogeneous property among source and target's head size and hairstyle must be solved.
When $I_s$ and $I_t$ have significantly different head regions, directly placing the head of $I_s$ within the masked region of $I_t$ can result in unnatural artifacts, as shown in ~\cref{fig:artifacts}.
Therefore, our goal is to prevent the model from inferring the head size and hairstyle only from the target image's mask contour.
Specifically, we randomly replace the conventional segmentation-based head mask with a more adaptive masking strategy designed to facilitate seamless head integration.
We apply various mask augmentation measures including dilation, widened bounding box creation, and merging with a random mask, as illustrated in~\cref{fig:mask}, similar to~\cite{selfswapper, choi2021viton}.

\subsection{Inference}
\label{subsec:infer}
\begin{figure}[t]
  \centering
   \includegraphics[width=1\linewidth]{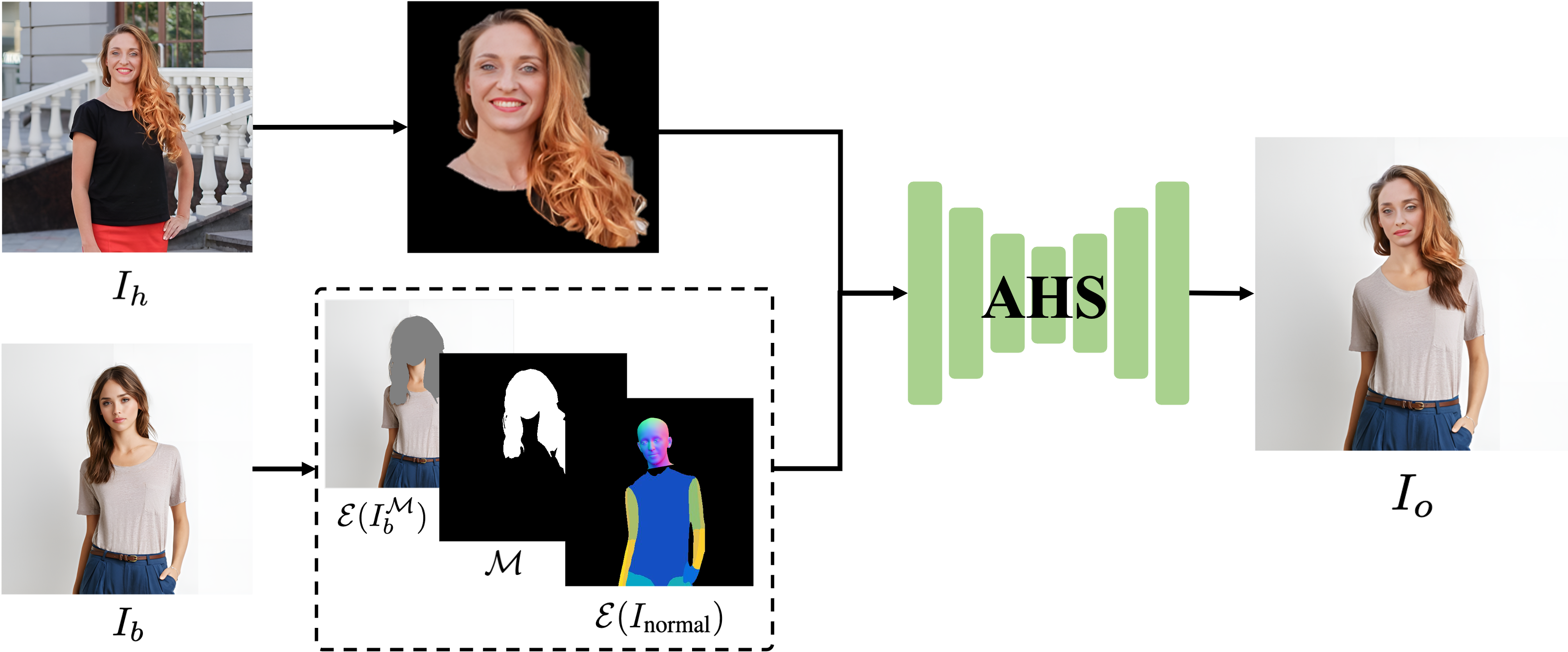}
   \caption{\textbf{Inference overview.} Our model takes source and target images and outputs head swapped results within an unified model.   }
   \label{fig:inference}
\end{figure}

Our proposed AHS enables efficient inference by leveraging its head reenactment-based training regime. Simply providing the target body and source head images as input allows the model to generate a head-reenacted and swapped result, as illustrated in \cref{fig:inference}. However, due to the discrepancies between input for training and inference, certain techniques are incorporated to ensure optimal performance.

\paragraph{Head parameter swapping.}
During training, since $I_s$ and $I_t$ are images of the same person, the learning can be accomplished using only the head normal map of $I_t$. However, during inference, the model is required to process images of two different individuals. 
Therefore, the shape parameters of $I_s$ must be reflected in the head normal map of $I_t$. 
To achieve this, after obtaining the FLAME parameters of both $I_s$ and $I_t$ through EMOCA~\cite{emoca}, we substitute the shape parameter of $I_s$ with the corresponding parameters of $I_t$ before generating the normal map.
\begin{equation}\label{eq:ca}
    \text{FLAME}(\boldsymbol{\beta}_{s}, \boldsymbol{\theta}_{t}, \boldsymbol{\psi}_{t}) \rightarrow (\textbf{V}, \textbf{F}),
\end{equation}
where FLAME~\cite{flame} is a statistical 3D head model that outputs vertices \textbf{V} and faces \textbf{F}, given the head shape parameter \textbf{$\beta$}, pose parameter \textbf{$\theta$}, and facial expression parameter \textbf{$\psi$}. The head normal map is subsequently rendered based on the extracted geometry information.

\paragraph{Inference mask.}
\label{sec:inference_mask}
When $I_s$ and $I_t$ differ significantly, especially for hair length and head size, performing inpainting using only the head mask of $I_t$ can lead to unnatural artifacts. For instance, the head may be generated too small or the hair may appear truncated. To address this, we first perform inpainting by providing a wide bounding box that covers a larger area than the original head mask of $I_t$. We then mask the head and neck portions of the generated image, and merge them with the mask of $I_t$ before performing inpainting once again. This approach increases the degree of freedom in generation while preserving the details outside the head area with minimal degradation in quality.

\section{Experiments}

\begin{table*}[t]
\centering
\resizebox{0.9\textwidth}{!}{
\begin{tabular}{l|cccccccc}
\hline
\textbf{Models}  & CLIP-I (Head) $\uparrow$   & FID  $\downarrow$   & FID (Crop) $\downarrow$& ID sim $\uparrow$  &Head orientation $\downarrow$ & Expression $\downarrow$ \\ 
\hline
\hline
REFace~\cite{reface}            & 0.7859             &         8.4491      &         24.31         & 0.5239             & \textbf{5.57}           & \underline{7.308}              \\ 
InstantID\textsuperscript{*}~\cite{instantid}    & 0.8223             &         5.7882      &         11.43         & 0.2829             & \underline{7.52}        & 7.591              \\ 
HID~\cite{ours}             & \underline{0.8577} &  \textbf{5.6306}    & \underline{6.83}      & \underline{0.5555} & 11.51                   & 8.474              \\ 
\hline
\rowcolor{gray!20}
AHS (Ours)                          & \textbf{0.9132}    & \underline{5.7818}  & \textbf{5.02}         & \textbf{0.6230}    & 8.01                    & \textbf{6.204}     \\
\hline
\hline
\end{tabular}
}
\caption{\textbf{Quantitative comparison.} Best and second-best results are in bold and underlined, respectively. Our proposed method, AHS, outperforms existing methods in most metrics, excluding FID and head orientation.}
\label{tab:quanti}
\end{table*}

To evaluate our method, we conduct experiments on the SHHQ~\cite{shhq} dataset. Following the experimental setup, we present quantitative and qualitative results with a user study to demonstrate state-of-the-art performance.
Finally, we provide a comprehensive ablation study with analysis. Further details can be found in~\cref{Appen:Imple}.

\subsection{Experimental Setup}
\label{sec:exp}

\paragraph{Implementation details.}
To enhance the output quality, we adopt an alternative scheduler~\cite{lin2024common} and unfroze the S-Net. The H-Net remains frozen as its pretrained features are already effective, and unfreezing it would incur prohibitive memory and time costs. Similarly, we froze the face encoder to preserve its pretrained weights ~\cite{photomaker} and unfroze the last layer the head encoder to enable residual detail enhancement from the input image.
Training is implemented with data type casting to bfloat16 to enhance efficiency and is conducted on 4 H100 80GB GPUs over 70 epochs, using a batch size of 6 per GPU, which takes about 3 days.

\begin{figure*}[t!]
  \centering
   \includegraphics[width=1.0\linewidth]{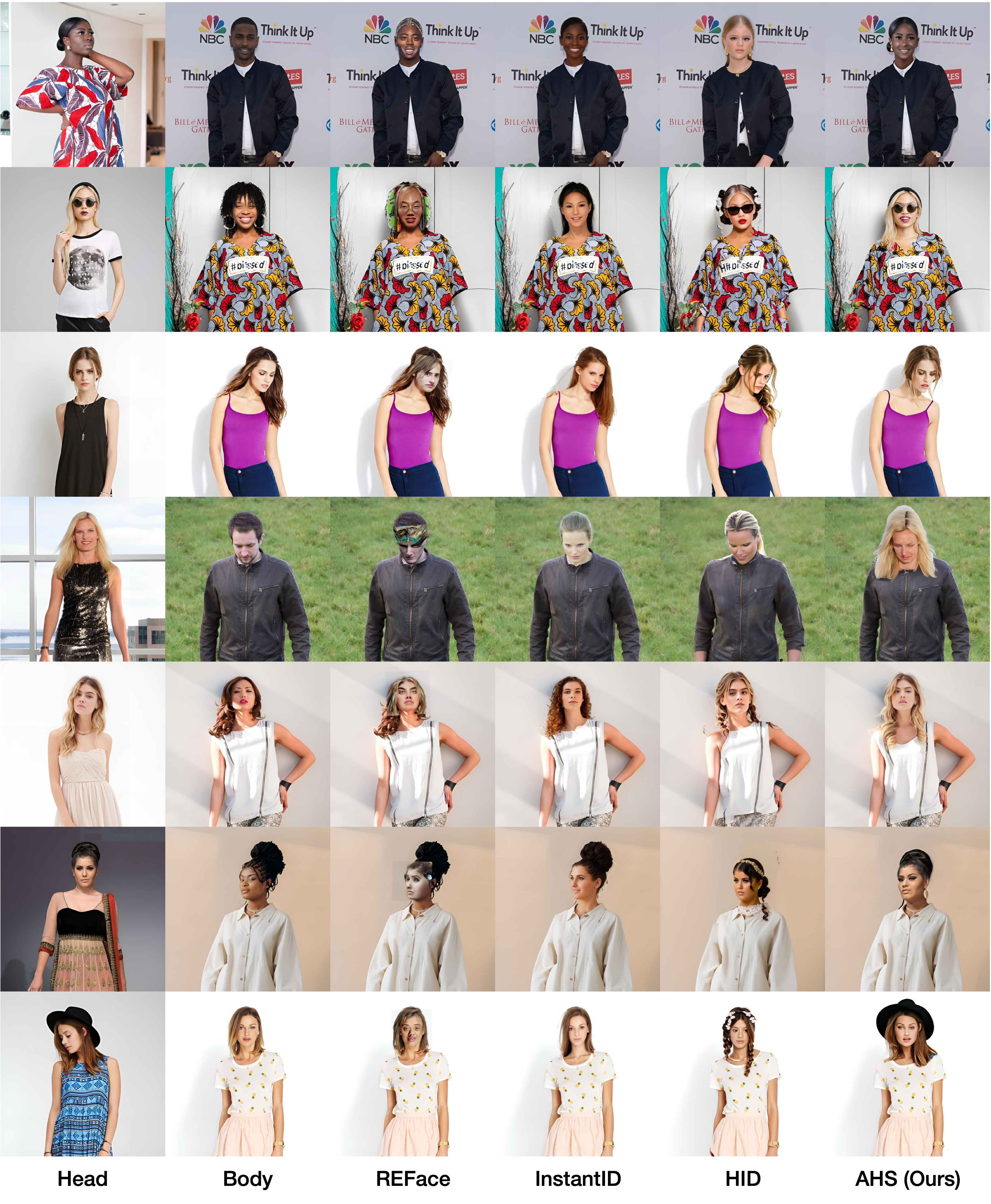}
   \caption{\textbf{Qualitative comparison.} 
    The images in the \textit{Head} column are combined with those in the \textit{Body} column. The last four columns are the head-swapped results produced by each method. More results can be found in \cref{Appen:Qual}.}
   \label{fig:qual_result}
\end{figure*}

\paragraph{Baselines.}
Due to the underexplored nature of the zero-shot head swapping task, there is a lack of established baselines for comparison. Among existing methods, HID~\cite{ours} is a zero-shot head swapping approach specifically designed for upper body datasets, aligning most closely with our task.
REFace~\cite{reface}, while not originally intended for head swapping, can also perform zero-shot head swapping. For a fair comparison, we utilize its weights trained for head swapping tasks.
Due to the scarcity of high-performing, dedicated head-swapping baselines, we chose to compare our method against InstantID~\cite{instantid}, a powerful and widely-used approach for identity-preserved generation. 
Although not originally designed for head swapping, incorporating IP-Adapter~\cite{ipadapter} and ControlNet~\cite{controlnet} with its architecture makes it adaptable for this task. We achieve this adaptation by leveraging the SDXL Inpainting model~\cite{podell2023sdxl} with Sapiens masks~\cite{khirodkar2024sapiens}. This setup allows the model to inpaint the head region to transfer identity, while concurrently using ControlNet to enforce the desired pose.
Consequently, we compare our method against HID, REFace, and InstantID for comprehensive evaluation.

\paragraph{Evaluation.}

For evaluation, we create test pairs by randomly selecting a source (head) image and a target (body) image from the dataset. We assess our method using a suite of quantitative metrics, following the protocol of our baseline HID~\cite{ours}, alongside a user study.
The quantitative evaluation includes measuring the overall visual quality using the Fréchet Inception Distance (FID)~\cite{fid}. We also assess more specific perceptual aspects with CLIP-I~\cite{clip} similarity measured exclusively on the generated head region where the face and hair areas are segmented by SCHP~\cite{schp} to isolate the quality. We further evaluate identity similarity with ArcFace~\cite{arcface}, head orientation error with HopeNet~\cite{hopenet}, and expression similarity with FLAME~\cite{flame}. To ensure a fair comparison against baselines that occasionally fail, the metrics are reported only on the common subset of samples where results are successfully generated across all methods. Consequently, scores for our more robust model may differ slightly between the main comparison and our ablation studies, where this filtering is not applied.

\begin{table*}[t]
\centering
\begin{minipage}[h]{0.35\textwidth}
  \centering
    \includegraphics[width=1.0\linewidth]{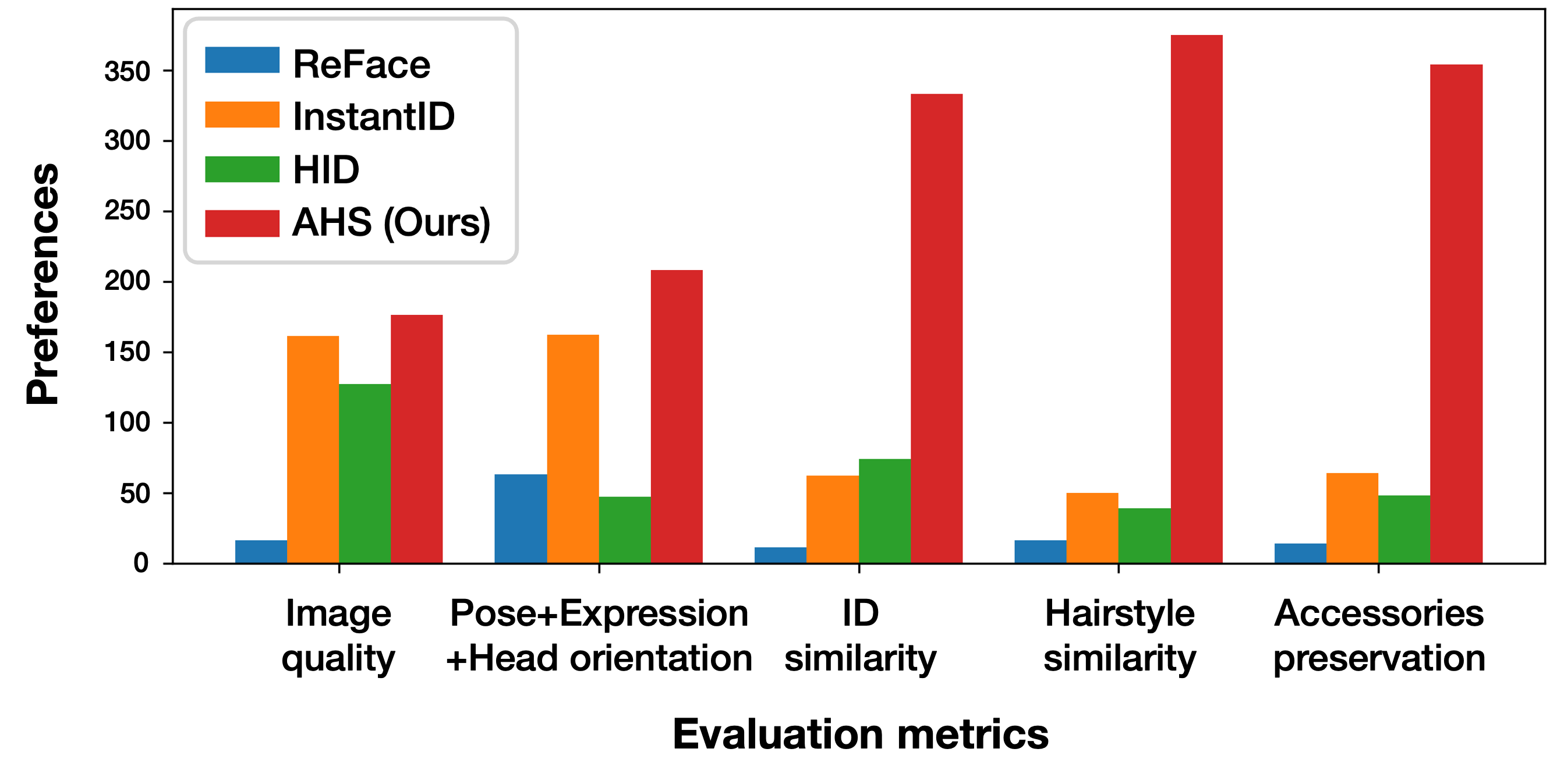}
     \captionof{figure}{\textbf{Results of user study.}}
    \label{fig:appen_user_study}
\end{minipage}
\begin{minipage}[]{0.63\textwidth}
\resizebox{\textwidth}{!}{
\begin{tabular}{l|cccccccc}
\hline
\textbf{Models}  & CLIP-I (Head) $\uparrow$ & FID $\downarrow$ & ID sim $\uparrow$  &Head orientation $\downarrow$ &  Expression $\downarrow$ \\ 
\hline
w/o GAG Aug             & 0.7575                    & 10.94              &  \textbf{0.8492}      &  17.34                      &  6.8542       \\ 
w/o Mask Aug            & 0.8637                    & \underline{8.99}   &  0.4098               & \textbf{7.27}               & \textbf{5.6719}   \\ 
\hline
w/o head encoder        & 0.8912                    & 9.29               & 0.5690                & \underline{7.90}            & 8.8247  \\ 
w/o face, head encoder  &  \underline{0.9019}       & 9.66               & \underline{0.6411}    & 8.92                        & 8.4452  \\ 
w/o decoupled CA        & 0.8050                    & 9.04               & 0.5124                & 8.27                        & 8.7458 \\ 
\hline
\rowcolor{gray!20}
AHS (Ours)                &  \textbf{0.9122}          & \textbf{8.98}      & 0.6237                & 8.72                        & \underline{6.4273}\\ 
\hline
\hline

\end{tabular}
}
\caption{\textbf{Ablation study quantitative results.} For each metric, the best and second-best performing results are denoted in bold and underlined formats, respectively.}
\label{tab:ab_st}
\end{minipage}

\end{table*}

\begin{figure}[t!]
  \centering
   \includegraphics[width=0.95\linewidth]{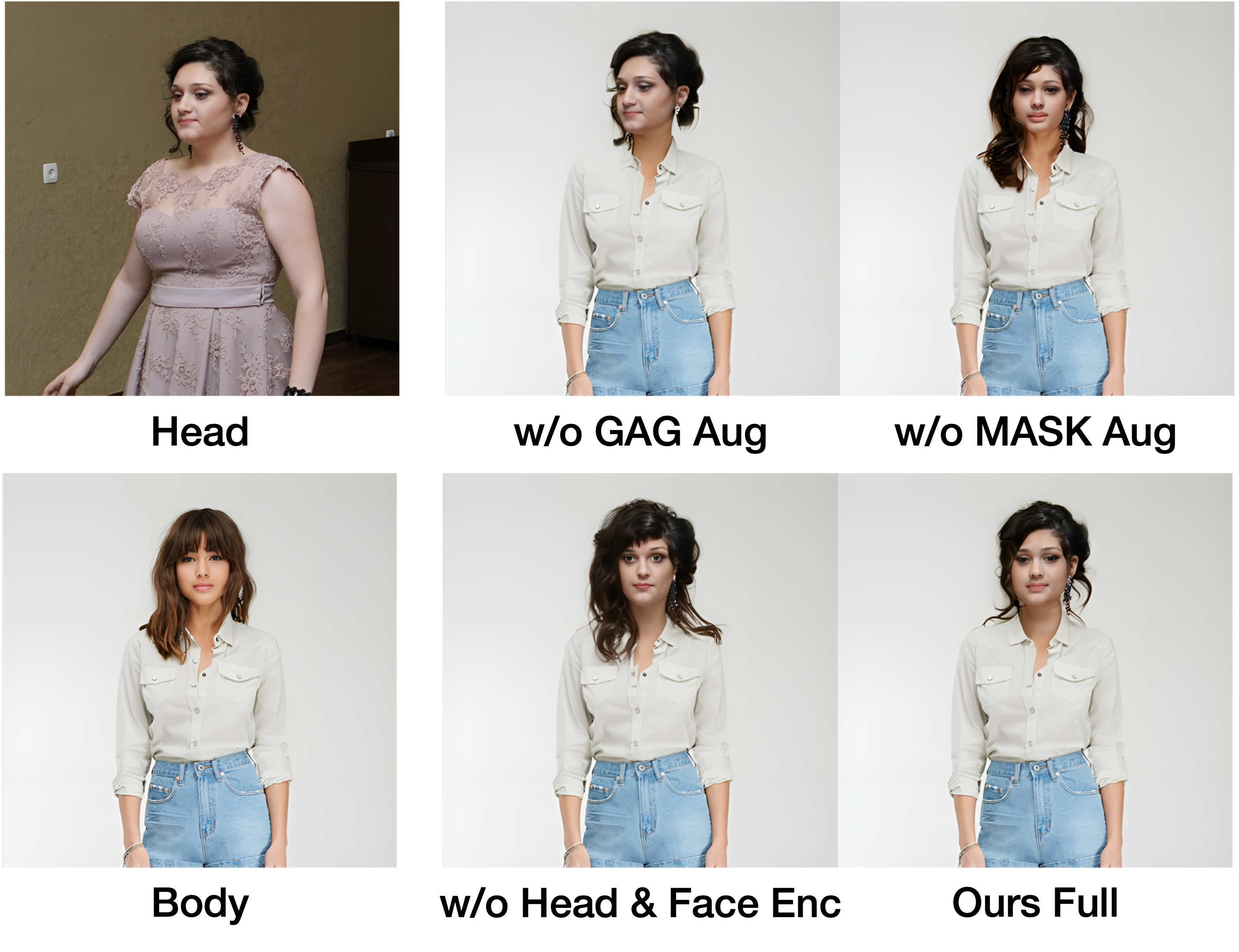}
    \caption{\textbf{Ablation study qualitative results.} Without the head and face encoders, identity preservation degrades under large head pose differences. Without augmentation, self-supervised learning causes copy-and-paste artifacts.}
    \label{fig:ablation}
\end{figure}

\subsection{Results}

\paragraph{Quantitative results.}
As shown in \cref{tab:quanti}, our proposed method, AHS, outperforms other methods across most metrics, achieving the best performance in identity similarity, FID (cropped), CLIP-I, and expression preservation.
While methods like REFace~\cite{reface} maintain head orientation well by directly inputting cropped faces and landmarks, they exhibit significant trade-offs, showing lower quality, identity similarity and expression. Similarly, InstantID~\cite{instantid} relies on a large ControlNet model, namely Identity Net, to inject head orientation information with landmark. In contrast, AHS achieves comparable orientation control simply by adding normal semantics to the input, demonstrating superior efficiency.
Ultimately, although REFace scores well on full-image FID due to its crop-and-paste nature, our model, AHS, which generates the entire image holistically, not only achieves comparable results in those areas but also demonstrates superior overall performance.

\paragraph{Qualitative results.} 
Unlike baseline methods, AHS achieves high identity preservation while maintaining facial expressions.
Baseline methods often suffer from a trade-off, where preserving expressions leads to identity degradation, or ensuring identity results in poor expression integration, as shown in~\cref{fig:qual_result} and~\cref{fig:teaser}.
In contrast, AHS successfully maintains identity while accurately reflecting the expressions of $I_t$. 
Additionally, leveraging $I_{normal}$ input and mask augmentation training enables AHS to outperform other methods in cases where there are significant pose variations or hairstyle differences.
While HID~\cite{ours} demonstrates strong identity preservation under extreme conditions compared to existing methods, AHS achieves superior performance by more robustly retaining identity and hairstyle, faithfully transferring the target's facial expressions, and robustly preserving accessories such as hats and sunglasses, even in highly challenging scenarios. We provide additional qualitative examples in~\cref{Appen:Qual} to further validate the robustness of the proposed approach.

\paragraph{User Study.}
To validate the effectiveness of our method, AHS, we conduct a multiple-alternative forced-choice user study on 20 image pairs involving 19 participants. Each participant is asked to select the better result based on several key aspects: overall image quality, reenactment fidelity (pose, expression, and head orientation), hairstyle similarity and accessories preservation. As summarized in Fig.~\ref{fig:appen_user_study}, AHS is consistently preferred across all criteria. It demonstrates a particularly significant lead in preserving identity features, as well as hairstyle and accessories.

\subsection{Ablation Study}

We perform an ablation study to analyze the impact of our key model architecture components (face encoder, head encoder, and decoupled CA) and data augmentation strategies (GAGAvatar~\cite{GAGAvatar} and mask). The results are presented in \Cref{tab:ab_st}.
First, we examine the model architecture. When both the face and head encoders are removed, the model's performance relies heavily on the S-Net. While this leads to a high ID-sim score, the lack of detailed structural guidance yields a significant drop in FID and a failure to transfer expressions. As seen in \Cref{fig:ablation}, this variant struggles to preserve identity, especially when handling large differences in head orientation and expression. Furthermore, removing the head encoder or the decoupled CA individually results in lower scores across most metrics, demonstrating that each component is necessary for optimal performance.
Next, we evaluate the augmentation strategies. The model trained without the GAGAvatar augmentation achieves a high ID-sim score, but it learns a "shortcut" by directly filling the masked area with conditional features. This approach results in prominent copy-and-paste artifacts, as shown in \Cref{fig:ablation}, and indicates a failure to properly learn the reenactment task. 
Without mask augmentation, the model fails to learn background inpainting and generates images following the hair silhouette in the source image.
These results reveal various trade-offs within the ablated models. However, they also validate that our final model achieves consistently superior performance across all metrics. This confirms that every component, the dual encoders for comprehensive guidance and our augmentation strategy for effective learning, is indispensable.

\section{Conclusion}

In this paper, we propose a novel head swapping approach, AHS, which effectively handles diverse head orientations, expressions, and hairstyles. AHS leverages a synthetic data augmentation strategy and a comprehensive conditioning approach using both cross- and self-attention. Experimental results demonstrate that AHS outperforms existing methods in identity preservation, expression transfer, and visual quality. 
However, head reenactment data augmentation using GAGAvatar~\cite{GAGAvatar} alone is insufficient to handle lighting variations. To address this, applying the same data augmentation strategy during the training using relighting models such as IC-Light~\cite{ic-light} mitigates sensitivity to lighting changes. Detailed experiments are shown in~\cref{Appen:ic-light}. 
Despite these improvements, some limitations still remain, particularly in edge cases involving extreme facial structure differences or severe occlusions. These scenarios may lead to visual artifacts as the model struggles with geometric mismatches or missing information. Therefore, future work will focus on improving the robustness of the model for broader real-world applications. 

\clearpage
\section*{Acknowledgments}
This work was supported by Institute for Information \& communications Technology Planning \& Evaluation(IITP) grant funded by the Korea government(MSIT) (RS-2019-II190075, Artificial Intelligence Graduate School Program(KAIST)) and the National Research Foundation of Korea(NRF) grant funded by the Korea government(MSIT) (No. RS-2025-00555621).

{
    \small
    \bibliographystyle{ieeenat_fullname}
    \bibliography{main}
}

\clearpage

\section{Additional Related Work}
\label{Appen:Related}
\subsection{Head Swap}
Many existing methods, such as those proposed in~\cite{deepfacelab, faceX, reface}, optimize their approaches based on these cropped datasets, leading to inherent limitations in handling cases where the full head or surrounding region should be harmonized with the body.
Consequently, these methods struggle with occlusions, head orientations beyond a narrow frontal distribution, and varying hair structures. 
While FaceX~\cite{faceX} and REFace~\cite{reface} leverage diffusion models for head swapping, they still rely on face-centered training data, inheriting the same dataset-induced weaknesses.
HSDiffusion~\cite{hsdiffusion}, although diffusion-based, assumes a simple alignment mechanism where the center points of the head and body images are matched before compositing. However, without explicit modeling of head orientation differences, this approach often results in unnatural compositions when the source and target images have misaligned orientations.
In contrast, HeSer~\cite{heser} attempts to address these limitations by incorporating more varied head orientations. However, it operates under a few-shot learning paradigm, making it less flexible and scalable compared to zero-shot approaches.
Additionally, the recent method GHOST 2.0~\cite{ghost} adopts HeSer’s blending technique, requiring precise image alignment similar to HeSer. This introduces more complex data preprocessing steps. Furthermore, due to dataset limitations, it struggles to handle cases where the subject has extremely long hair.

\section{Implementation Details}
\label{Appen:Imple}

Our model is composed of three key components: the H-Net, which utilizes an SDXL inpainting model~\cite{diffusers_sdxl_inpainting_0.1}; the S-Net, which employs the UNet from the original SDXL~\cite{podell2023sdxl}; and a pretrained IP-Adapter~\cite{ipadapter} and a pretrained face encoder from PhotoMaker~\cite{photomaker}. We train our model on the SHHQ dataset~\cite{shhq}, adopting the data handling procedures from HID~\cite{ours} with modified captions as detailed in \cref{Appen:dataset}. For data augmentation, we apply GAGAvatar~\cite{GAGAvatar} to 70\% of the images. The corresponding masks are augmented through dilation (with a 90\% probability), concatenation (50\%), and conversion to bounding boxes (50\%). The model is trained for 70 epochs using the AdamW optimizer~\cite{adamw} with a learning rate of $1 \times 10^{-5}$ and a batch size of 6 per GPU. For reproducibility, we fix the random seed to 42 for both training and inference. During inference, we use a classifier-free guidance (CFG)~\cite{cfg} scale of 2.0 with 30 denoising steps and generate images at a resolution of $1024 \times 1024$. In addition, for simplicity and efficiency, we employ the DeepXception model~\cite{chollet2017xception} to generate the segmentation mask during inference.


\subsection{Datasets}
\label{Appen:dataset}
We leverage the SHHQ dataset~\cite{shhq} following the approach in HID~\cite{ours}, but modify the captions. By replacing the original text embedding of 'hairstyle' with a fused embedding from the hair image and text 'hairstyle', HID eliminates the need for hairstyle descriptions in the prompt.Instead, we generate image captions about the hairstyle using the multi-modal large language model, GPT-4o~\cite{gpt4} and add the generated captions to each original caption used in HID after removing "with hairstyle". To explicitly indicate the hair region, we provide both the input image and the cropped hair portion of the input image to the model. 


\subsection{Loss}
To train our model, we use a composite loss function that balances overall image fidelity with accuracy in the specific head region. Our final loss, $\mathcal{L}_{\text{total}}$, is formulated as a weighted sum of two Mean Squared Error terms:
\begin{equation}
    \mathcal{L}_{\text{total}} = \lambda_1 \mathcal{L}_{\text{global}} + \lambda_2 \mathcal{L}_{\text{head}}
\end{equation}
Here, $\mathcal{L}_{\text{global}}$ is the standard MSE loss between the prediction and the ground truth over the entire image. $\mathcal{L}_{\text{head}}$ is an MSE loss computed exclusively on the head region, isolated using a mask $M$. For all experiments, we set the balancing hyperparameters $\lambda_1$ and $\lambda_2$ to 1.0.

\subsection{GAGAvatar Augmentation}
For GAGAvatar~\cite{GAGAvatar} augmentation, we employ a balanced sampling strategy to ensure robustness.
Head pose differences are distributed as $5^{\circ}$ (37\%), $5\text{--}10^{\circ}$ (31\%), and $>15^{\circ}$ (32\%). For expression variations, we sample across a wide range of $[-0.52, 0.99]$, achieving a mean cosine similarity of $0.67$ ($\sigma=0.18$). This diverse distribution allows the model to generalize across various motion scales. Our pipeline is fully automatic and does not require manual alignment between the source and target. Since the model generates the head within the target bounding box while centering the head using a normal map, it remains robust to large pose disparities.

\begin{figure*}[t!] 
    \centering
    
    \includegraphics[width=0.9\linewidth]{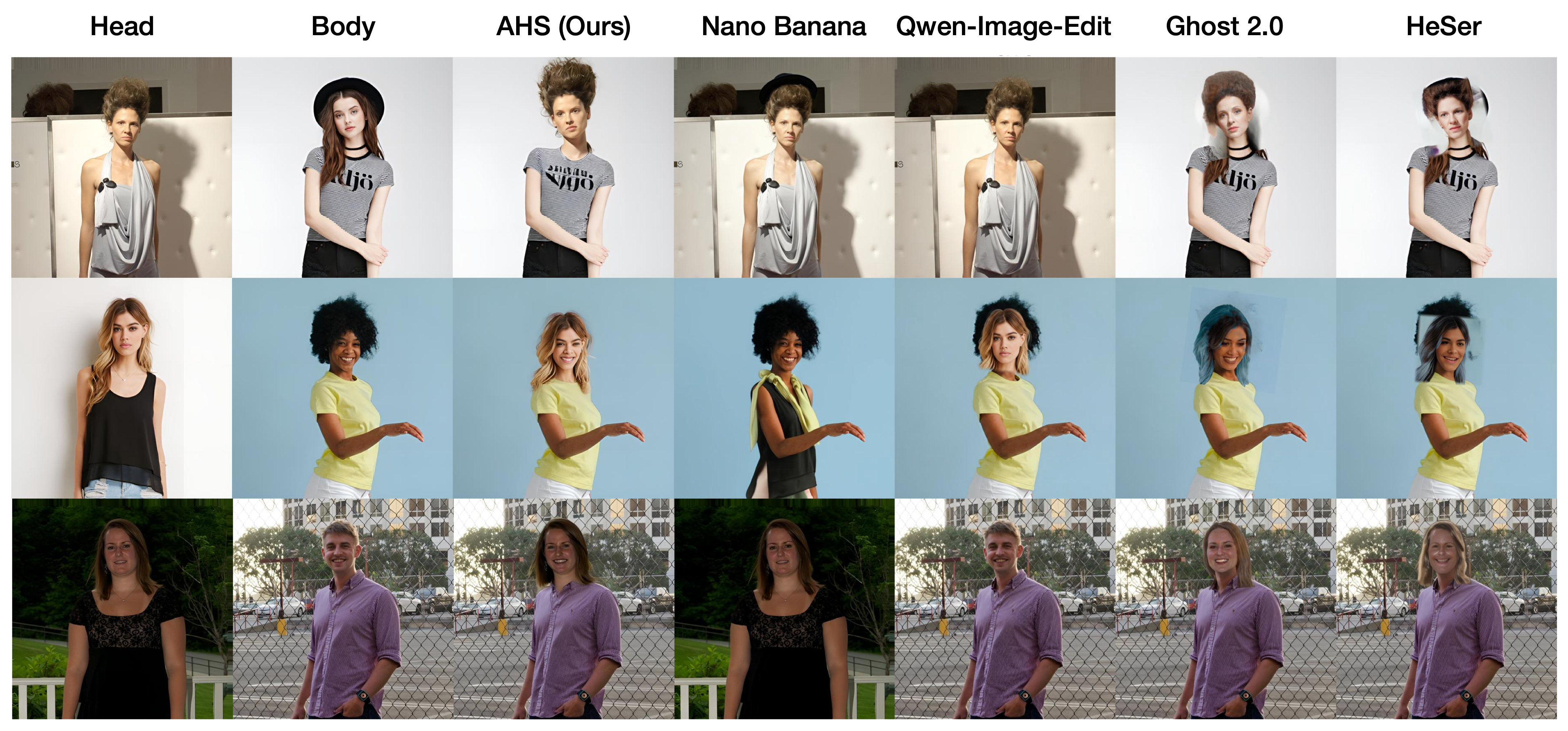}
   \caption{\textbf{Qualitative comparison with additional baselines.}}
   \label{fig:r1}
\vspace{16pt}

  \includegraphics[width=0.75\linewidth]{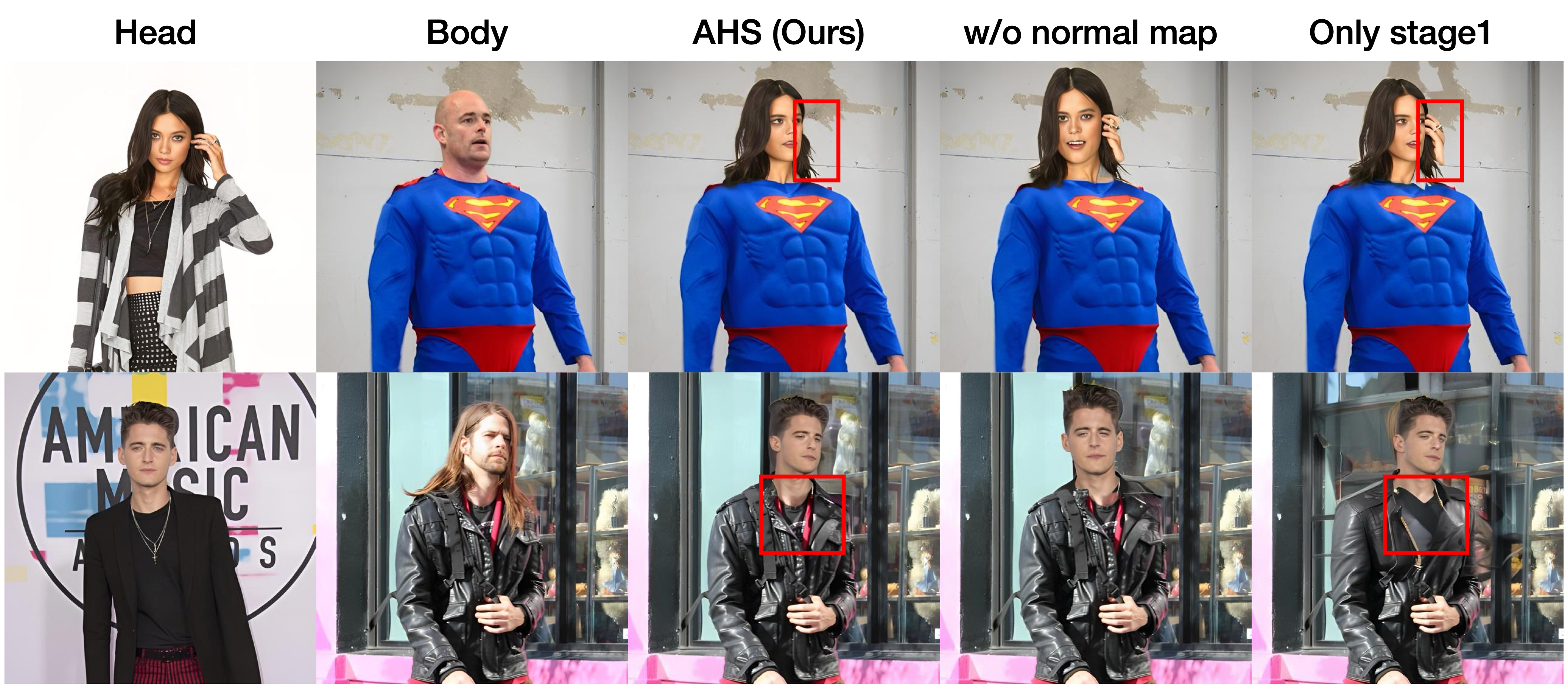}
   \caption{\textbf{Qualitative results of additional ablation study.}}
   \label{fig:r2}

\vspace{16pt}

\resizebox{0.7\linewidth}{!}{
\captionsetup{type=table}
\begin{tabular}{l|ccccccc}
\toprule
Method            & CLIP-I(Head) $\uparrow$ & FID $\downarrow$ & FID(Crop)  $\downarrow$ & ID sim$\uparrow$  & Head orientation$\downarrow$  & Expression $\downarrow$\\
\drule 
Ghost 2.0           & 0.8487            & 7.968             & 19.749            & 0.483               & \textbf{4.831}      & 10.626            \\
HeSer~\cite{heser}  & 0.8507            & 6.444             & 15.221            & 0.503               & 8.734               & 11.022            \\
Nano Banana~\cite{nanobanana}     & 0.8634            & \textbf{4.241}    & \textbf{2.809}    & 0.474               & 10.725              & 7.449             \\
Qwen-Image-Edit~\cite{qwen}              & 0.8789            & \underline{4.377} & \underline{4.422} & 0.536               & 15.504              & 7.522             \\
\drule
Ours                & 0.9139            & 9.613             & 6.719             & 0.625               & \underline{8.427}   & \underline{6.887} \\
\drule
Ours w/o normal     & \textbf{0.9238}   & 10.055            & 5.944             & \textbf{0.729}      & 17.472              & 10.799            \\
Ours stage1         & \underline{0.9157}& 9.073             & 6.221             & \underline{0.631}   & 9.000               & \textbf{6.761}    \\
\bottomrule
\end{tabular}
}
\captionof{table}{\textbf{Quantitative Results.} }
\label{tab:r1}

\end{figure*}

\section{Additional Experiments}

\subsection{Comparisons with Additional Baselines}
We further compare our AHS with four additional baselines: Nano Banana~\cite{nanobanana}, Qwen-Image-Edit~\cite{qwen}, HeSer~\cite{heser}, and Ghost 2.0~\cite{ghost}. As shown in \cref{fig:r1} and \cref{tab:r1}, while Nano Banana and Qwen-Image-Edit prioritize consistency, they often produce images identical to the input or suffer from severe copy-and-paste artifacts, which paradoxically leads to a high FID score.
Similar to REFace~\cite{reface}, both HeSer and Ghost 2.0 rely on a conventional face-swap paradigm based on a crop-and-align pipeline. This approach is inherently unsuitable for head swapping as it is unable to handle regions outside the fixed facial crop, such as long hair. As a result, all three methods suffer from prominent bounding box artifacts and degradation in structural completeness compared to our method.

\subsection{Additional Ablation Study}
As shown in \cref{fig:r2} and \cref{tab:r1}, omitting surface normals degrades head orientation and expression accuracy, as they provide essential geometric guidance for motion-identity decoupling. Furthermore, while bounding-box-only inference lacks boundary constraints leading to background flickering and deformation.

\subsection{Lighting Condition Augmentation}
\label{Appen:ic-light}
\begin{figure}[t]
  \centering
   \includegraphics[width=1.0\linewidth]{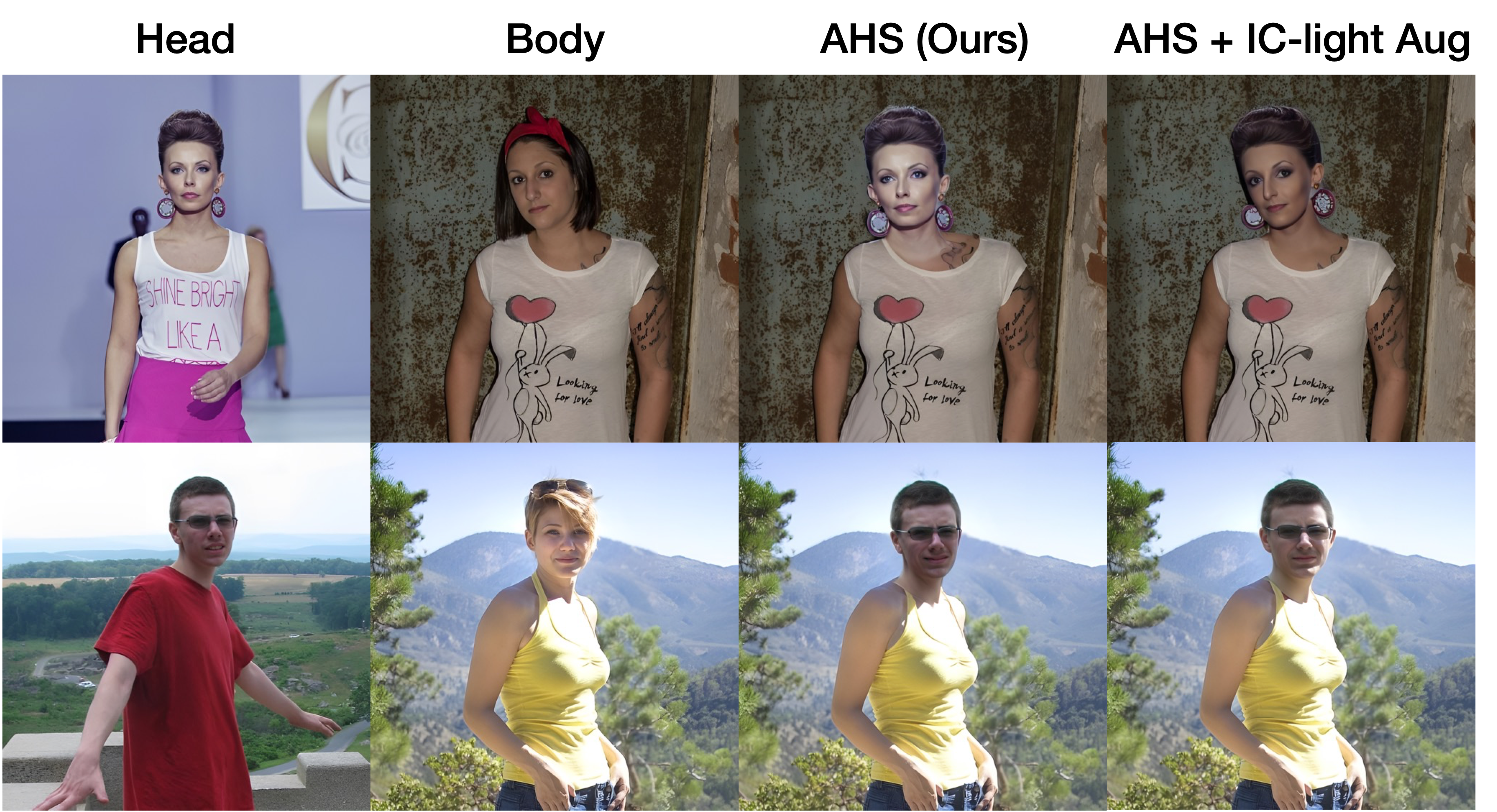}
   \caption{\textbf{Results of IC-light Augmentation.}}
   \label{fig:appen_ic-light}
\end{figure}

Regarding lighting variations and complex occlusions, we realize that GAGAvatar alone is insufficient to handle these factors. To address this, we apply a data augmentation strategy using relighting models such as IC-Light~\cite{ic-light}. Specifically, during training, we further augment 70\% of the images augmented by GAGAvatar by applying IC-Light. As qualitatively shown in~\cref{fig:appen_ic-light}, this approach mitigates sensitivity to lighting changes and further enhances overall lighting consistency.


\subsection{Inference Mask}
\begin{figure}[t]
  \centering
   \includegraphics[width=1.0\linewidth]{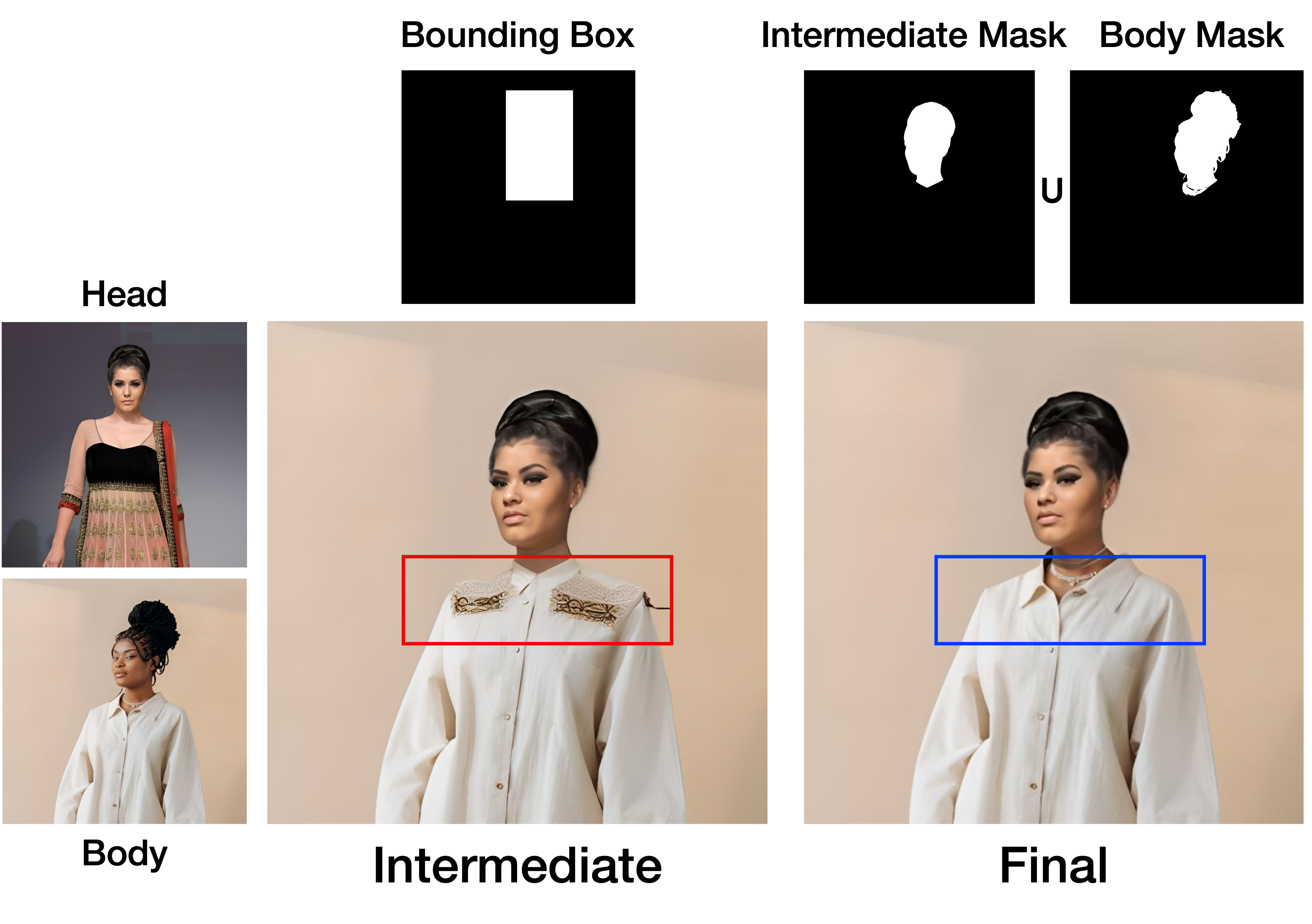}
   \caption{\textbf{Inference mask results.}}
   \label{fig:appen_infer_mask}
\end{figure}

As detailed in Section~\ref{sec:inference_mask}, this section elaborates on our inference methodology.
Generating realistic hair presents a unique challenge, as the target area requires significant creative flexibility. To accommodate this, we employ a two-step inference process.
Initially, we perform inference using a simple bounding box as the mask. While this approach provides ample space for hair synthesis, its broad nature can lead to undesirable artifacts, such as the deformation of clothing or the background outside the primary head region.
To address this, we refine the mask in a second step. First, we extract a precise head region mask from the intermediate output. We then create a new, more accurate mask by computing the union of this mask and a mask from the body image. This refined mask is used to perform a second round of inference. This strategy ensures that the inpainting process is precisely focused on the desired areas, preventing modifications to irrelevant regions.
As illustrated in \Cref{fig:appen_infer_mask}, the intermediate result, while plausible, exhibits clothing distortion. In contrast, the final output is cleanly reconstructed because the refined mask correctly excludes the clothing area from the inpainting process.

\section{Failure Cases}
\begin{figure}[t]
\centering
  \centering
  \includegraphics[width=1.0\linewidth]{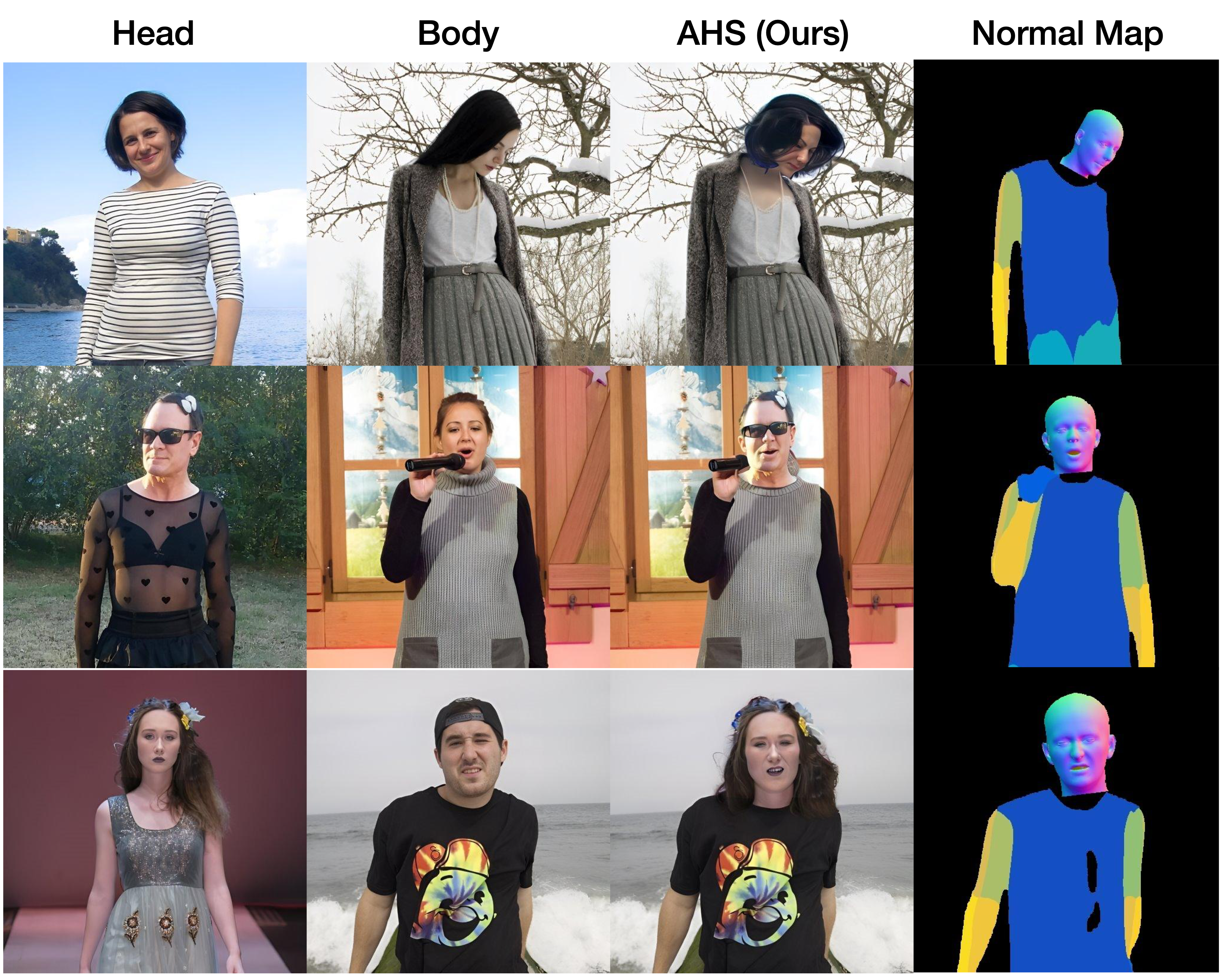}
   \caption{\textbf{Failure Cases.}}
   \label{fig:r3}
\end{figure}

Despite robust normal estimation in profile views, \cref{fig:r3}, our method faces three main challenges: (1) identity preservation under extreme poses, (2) restoration of masked-out facial occlusions, and (3) maintaining consistent facial scales when aligning with the body geometry. These cases arise from the inherent difficulty of hallucinating out-of-distribution spatial and structural information.

\begin{figure*}[t]
  \centering
   \includegraphics[width=1.0\linewidth]{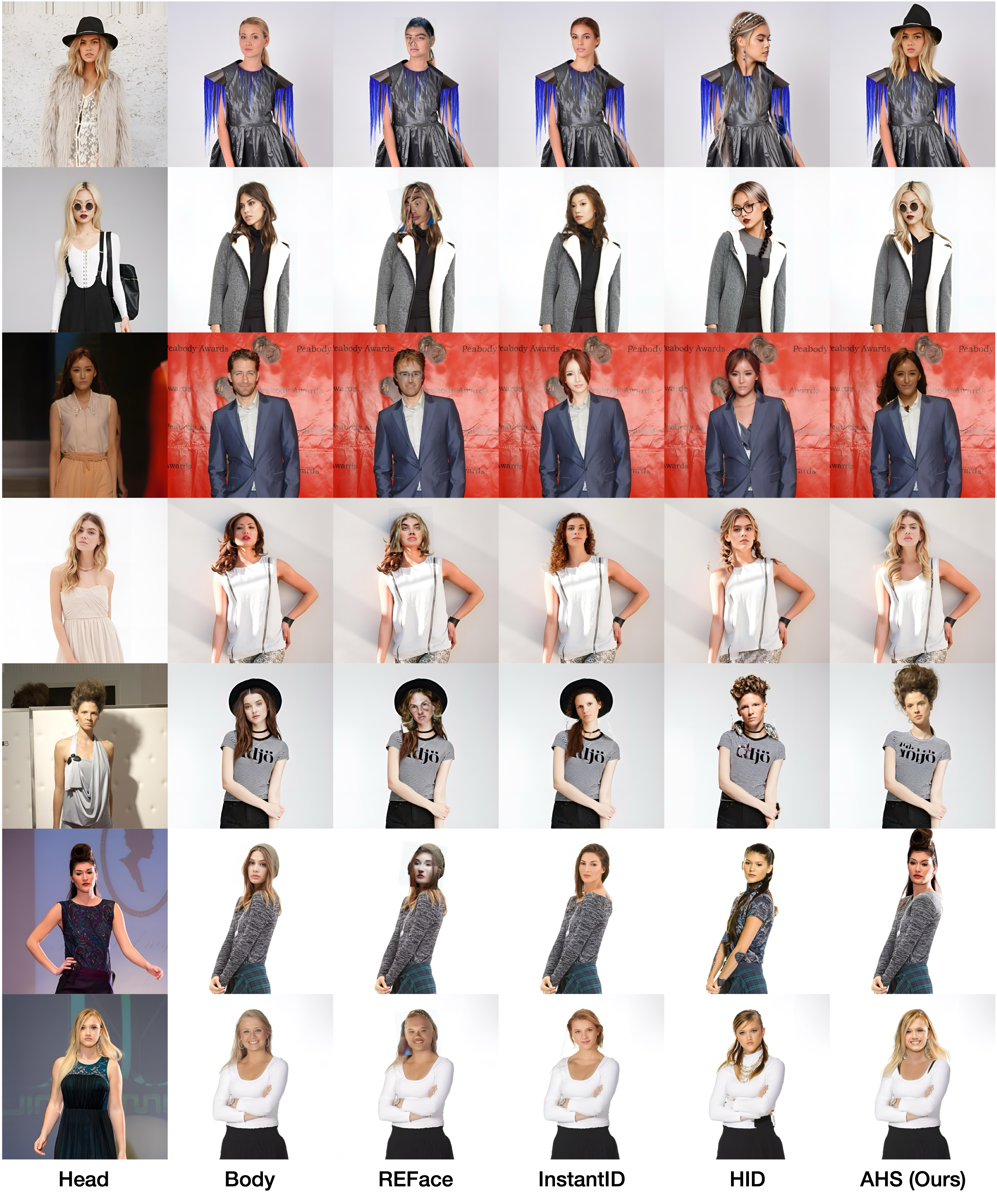}
   \caption{\textbf{Qualitative comparison.} 
    The images in the \textit{Head} column are combined with those in the \textit{Body} column. The last four columns are the head-swapped results produced by each method.}
   \label{fig:appen_qual_result_1}
\end{figure*}

\begin{figure*}[t]
  \centering
   \includegraphics[width=1.0\linewidth]{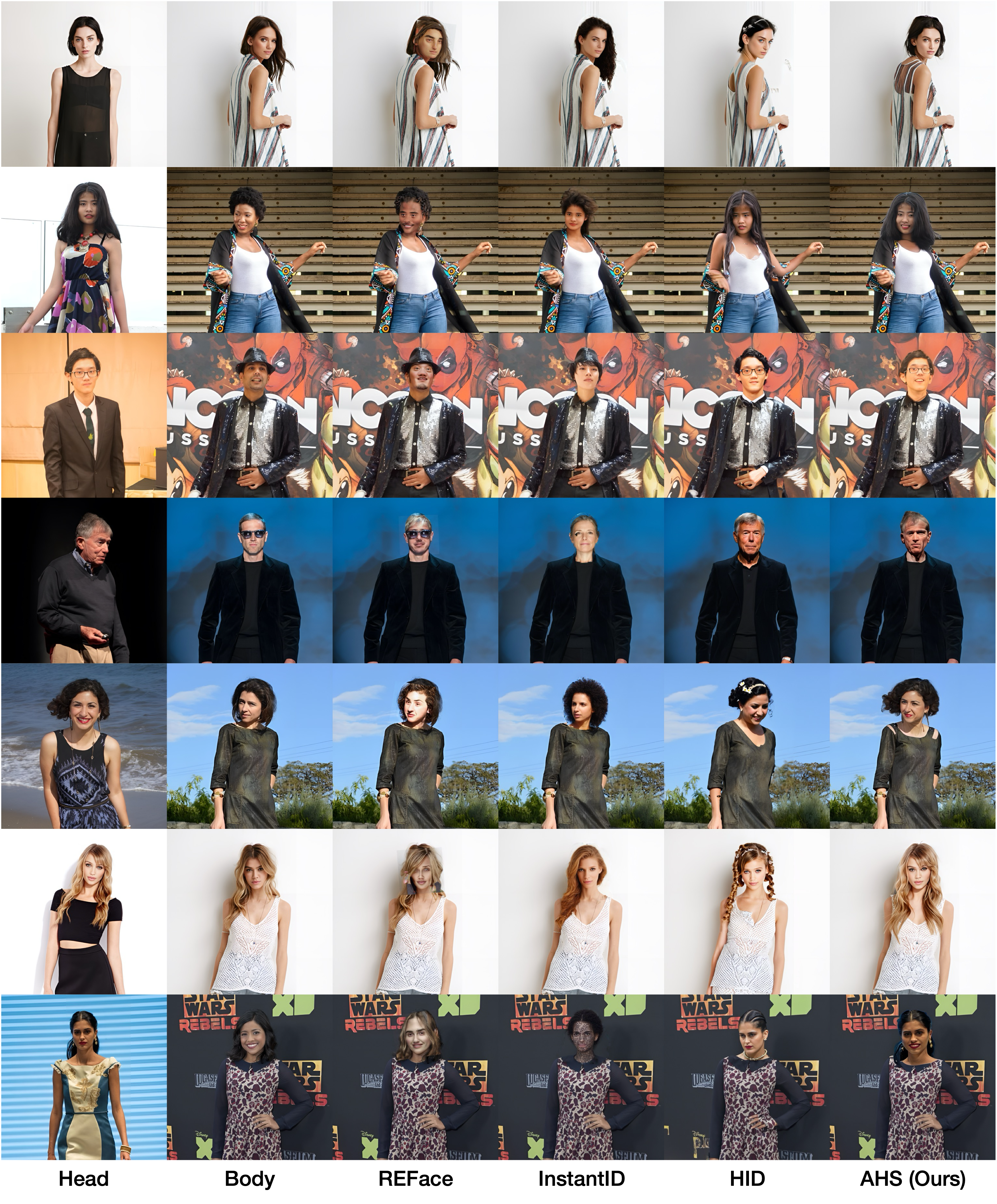}
   \caption{\textbf{Qualitative comparison.} 
    The images in the \textit{Head} column are combined with those in the \textit{Body} column. The last four columns are the head-swapped results produced by each method.}
   \label{fig:appen_qual_result_2}
\end{figure*}

\section{Additional Qualitative Results}
\label{Appen:Qual}
We provide additional qualitative results in \cref{fig:appen_qual_result_1} and \cref{fig:appen_qual_result_2} generated by our proposed approach.

\end{document}